\documentclass{article} %
\usepackage[preprint]{colm2025_conference}

\usepackage{microtype}
\usepackage{hyperref}
\usepackage{url}
\usepackage{booktabs}

\usepackage{lineno}

\usepackage{xcolor}
\usepackage{colortbl}
\usepackage{amsmath}
\usepackage{graphicx}
\usepackage{subfigure}
\usepackage{subcaption}
\usepackage{multirow}
\usepackage{booktabs}
\usepackage{caption}
\usepackage{tabularx}
\usepackage{enumitem} 

\definecolor{darkblue}{rgb}{0, 0, 0.5}
\hypersetup{colorlinks=true, citecolor=darkblue, linkcolor=darkblue, urlcolor=darkblue}

\title{Steering Large Language Models with Register Analysis \\ for Arbitrary Style Transfer}

\author{Xinchen Yang \\
Department of Computer Science \\
University of Maryland, College Park \\
\texttt{xcyang@cs.umd.edu} \\
\And
Marine Carpuat \\
Department of Computer Science, UMIACS \\
University of Maryland, College Park \\
\texttt{marine@cs.umd.edu}
}

\begin{document}

\ifcolmsubmission
\linenumbers
\fi

\maketitle

\begin{abstract}
Large Language Models (LLMs) have demonstrated strong capabilities in rewriting text across various styles. However, effectively leveraging this ability for example-based arbitrary style transfer—where an input text is rewritten to match the style of a given exemplar—remains an open challenge. A key question is how to describe the style of the exemplar to guide LLMs toward high-quality rewrites. In this work, we propose a prompting method based on register analysis to guide LLMs to perform this task. Empirical evaluations across multiple style transfer tasks show that our prompting approach enhances style transfer strength while preserving meaning more effectively than existing prompting strategies.

\end{abstract}

\section{Introduction}

Text style transfer (TST) refers to the task of transforming an input text into a target style (e.g. formality) while preserving non-style attributes such as meaning and fluency ~\citep{mir-etal-2019-evaluating, strap, ST_survey_jin}. TST has many downstream applications. For example, one application is the intelligent writing assistant, which helps rewrite texts to meet users’ personalized requests (e.g. more professional, polite, etc.) \citep{ST_survey_jin}. Other applications include text simplification, text detoxification, authorship obfuscation and so on \citep{ST_survey_jin, detoxification, fisher2024styleremix}. 

Recent advances in natural language generation (NLG) and large language models (LLMs) have made it possible to perform TST tasks automatically at scale \citep{ST_survey_jin}. In response to a growing need for TST methods with reduced data requirements and broader style coverage \citep{ST_survey_jin, hu2023textstyletransferreview}, research community has increasingly focused on a general formulation of style transfer, \textit{arbitrary TST}, in which an LLM rewrites an input text into an arbitrary style specified by the user at inference-time \citep{PaR, recipe, styll}. \citet{recipe} shows promise in framing arbitrary TST as a sentence rewriting task by using natural language instructions such as "make this melodramatic". Despite of its flexibility, end-users are left with the task of constructing the right prompt for a desired style, which often requires opaque prompt engineering, and is particularly difficult for non-native speakers or other users  unfamiliar with how to express exact stylistic nuances in technical terms. Reacting to this, example-based arbitrary TST comes as a solution, where representative target-style exemplars are provided at inference-time, allowing LLM to infer the desired style from exemplars without requiring users to explicitly characterize it. For example, \citet{styll} introduces STYLL, an example-based abitrary TST method, where the LLM is prompted to summarize a few target-style exemplars provided at inference-time into a list of open-ended style descriptors (e.g. "clear, concise, persuasive, intelligent") before applying them to rewrite the input text. However, relying on such open-ended descriptions, whether user-provided or model-inferred, may poses challenges. While these descriptors may help move the text away from the source style, they do not always ensure faithful reproduction of the target style \citep{styll}. Moreover, as the style descriptors are \textit{unconstrained}, it is unclear whether they may have side effects such as muddying the line between style and content and thus causing unintended meaning alteration.

To address the above challenges, we propose prompting LLMs to analyze exemplars' style using Biber's multidimensional register analysis (MDA) framework. Our modeling hypothesis is that Biber’s register analysis provides a structured and effective way to generate accurate target style descriptors that LLMs can reliably use in TST generation. First, because Biber’s register analysis framework is widely available online and used in educational and linguistic contexts, LLMs are likely to have been exposed to examples of such analyses during training. Second, Biber’s approach is data-driven, grounded in empirical analysis of large corpora, with its key register dimensions found across multiple languages and in online texts \citep{biber_cross_lingual, biber_online}, suggesting that Biber's framework is useful to highlight style variations that are salient in LLM pre-training data, of which online texts constitute a large portion. As a result, using Biber's register analysis to describe examplars may produce theory-grounded, easy-to-follow style descriptors for LLMs. Additionally, we explore whether contrasting the style of input and target exemplars yield better results than characterizing the style of target exemplar \textit{only}.

In this work, we evaluate two prompting variants, with one of them based on Biber's register analysis and input-target style contrast while the other one based on register analysis only, against several baselines on a diverse range of style transfer tasks, including authorship imitation, formality transfer and text simplification. Empirical results show that our prompting approach enhances style transfer quality: (1) Our prompting variants show similar to improved style transfer strength compared to the baselines; (2) Our prompting variants show a large gain in meaning preservation across the tasks.

Our main contributions are as follows:
\begin{itemize}
    \item We propose a prompting method based on register analysis to enhance the quality of example-based arbitrary TST. In addition, we investigate the impact of input-target contrast v.s. characterizing target only across different use cases.
    \item Experiments across diverse style transfer tasks show that our approach enhances rewriting quality, with similar to better style transfer strength and remarkable gains in meaning preservation, suggesting better decoupling of style and content.
\end{itemize}

\section{Background}

\subsection{Style and Register in Corpus Linguistics}

In linguistics, \textit{style} refers to the language habits of one person (e.g. Shakespeare), or a group of people at one time or over a period of time (e.g. Old English "heroic" poetry) \citep{Crystal1969InvestigatingES}. Style reflects an individual's linguistic idiosyncrasies, and thus has applications in disputed authorship resolution, forensic linguistics and so on \citep{Crystal1969InvestigatingES, rudman2005non_disputed, forensic_linguistics}. \textit{Register}, on the other hand, refers to linguistic variation associated with the situational use of language and are generally described by three components: situational context, linguistic features (e.g. lexical and grammatical characteristics), and functional relationships between the first two \citep{biber, halliday1989language, biber2009register}. For example, registers can be characterized by speech / writing situation and communicative purposes (e.g. personal letter, academic, narrate, etc.) \citep{biber2009register}. In this view, linguistic features are always \textit{functional}: they tend to occur in a register because they are suited to the situational and communicative context of the register \citep{biber2009register}. \\

One framework to analyze authorship style is \textit{stylometry}. Typically, stylometric analysis involves statistical analysis of relative frequencies of common words, especially functional words (i.e. words with little lexical meaning and expressing grammatical relationships among other words) \citep{binongo2003wrote, argamon2018computational, jackgrieve}. Stylometric analysis has been successful in distinguishing authorship styles \citep{shakespeare2016new, taylor2017new}. However, it lacks explainability and backing of linguistic theories, making it insufficient in applications such as forensic investigation, where such requirements are expected for legal justifications \citep{jackgrieve}.  An alternative framework is \textit{register analysis}. \citet{jackgrieve} argued that "authors write in subtly different registers", showing that register analysis identifies the same underlying patterns of linguistic variation as stylometry. Thus, register analysis is a strong candidate framework to characterize style variation, as it can distinguish styles as effectively as stylometry and provides better explainability \citep{jackgrieve}.

\subsection{Style Transfer in NLP}

\paragraph{Style in NLP.} In NLP, style transfer tasks adopt a loose extension of the notion of linguistic style to general attributes in text, such as formality, sentiment, and so on \citep{ST_survey_jin}.  Some of these attributes are not strictly stylistic features from a linguistic perspective. For example, positive v.s. negative sentiment is arguably more of a content-related attribute than a stylistic one, and formality is more closely related to register than style. In practice, style distinctions are defined by the reliance on specific corpora, which limits LLMs' capability to adapt to a broad range of unseen styles and perform low-resource style transfer.

\paragraph{Supervised Single-Style Transfer} 
Supervised single-style transfer involves altering a specific stylistic attribute of text while preserving content, typically relying on large parallel style corpora for model training. 
For example, \citet{jhamtani2017shakespearizingmodernlanguageusing} introduced a copy-enriched Seq2Seq model to enhance content preservation for modern-Shakespearean style transfer, trained on a large parallel corpus curated by \citet{xu2012paraphrasing_shakespeare}. Recent advances in transformer-based models \citep{vaswani2017attention} have added new momentum. For example, \citet{de2021formalstyler} fine-tuned GPT-2 \citep{gpt2} on the GYAFC \citep{gyafc} formality transfer parallel corpus and observed improved performance. \citet{atwell2022appdia_offensive} introduced an offensive-inoffensive parallel corpus based on Reddit, and a discourse-aware mechanism to reduce offensiveness and preserve meaning.

\paragraph{Multitask Style Transfer.} 
Multitask style transfer aims to enable a single model to perform multiple style transfer tasks, allowing flexible style switching at inference-time. For example, one early work is the introduction of an unsupervised training framework which enables modifying multiple attributes of a text simultaneously while preserving content \citep{logeswaran2018content}. In another line of work, \citet{vecchi2022transferring} proposed a learning framework that separates the latent spaces of style and content, enabling multi-style transfer with improved content preservation. \citet{subramanian2018multiple}, on the other hand, indicated that disentangling style and non-style features is not necessary for multitask style transfer by showing that entangled models can work well under unsupervised or pseudo-supervised training. While multitask style transfer reduces reliance on parallel data by learning from target-style text, it still requires large amounts of non-parallel style-specific data — resources often unavailable for all but a few commonly studied styles.

\paragraph{LLM-Based Style Transfer.} Recent advances in techniques such as in-context learning \citep{gpt3} and instruction-tuning \citep{RLHF} have enabled LLMs to perform style transfer using only natural language prompts and in-context examples, without task-specific fine-tuning. One seminal work in this direction is \citet{recipe}, where style transfer is achieved by prompting general-purpose pre-trained LLMs (e.g. GPT-3) with rewriting instructions (e.g. "make this more comic"). Inspired by this, \citet{styll} designed an arbitrary TST method in a more fine-grained manner, which prompts LLMs to extract style descriptors from a few target-style exemplars as rewriting instructions. LLM-based style transfer greatly reduces the need for data and supervision, making it well-suited for low-resource style transfer. Despite its promises, this field of arbitrary TST remains relatively underexplored, with challenges in style controllability, content preservation, etc. In this work, we introduce a method based on register analysis to address these challenges.

\section{Approach}

In arbitrary TST, the target style is not defined by a predefined set of categories, but specified freely by the user. This makes the task more flexible but also more challenging: how can users convey a style that may be highly personal, nuanced, or unfamiliar to the model? Users may \textit{feel} about the style — ”it sounds like me”, or "I want the text to sound like this example", but find it hard to describe the stylistic complexities in explicit technical terms. Hence, we adopt a task formulation as following: \\
\textbf{Input:} an input text $x^{\mathrm{input}}$ and a style exemplar $x^{\mathrm{style}}$ indicating the desired style. 
\\\textbf{Output:} $x^{\mathrm{output}}$, a rewrite of $x^{\mathrm{input}}$ in the style of $x^{\mathrm{style}}$ while preserving meaning.

To approach this task, we start with the most straightforward method: prompting. Given that register analysis can be used as a basis to characterize, distinguish and explain authorship styles, we hypothesize that prompting LLMs with instructions based on register analysis enables more effective arbitrary TST. Here, we design a three-step prompting strategy to guide LLMs to perform example-based arbitrary TST. We present two variants: (1) \textit{RG-Contrastive} ("RG": abbreviation of "register"): prompting with "register analysis" + "contrasting input and output exemplars"; (2) \textit{RG}: an ablation of with the first variant with "register analysis" only. The pipeline is shown in Fig~\ref{fig:prompting_pipeline}. For full prompts used in the experiments, see Table~\ref{tab:full_prompts} in Appendix~\ref{full_prompts}. 

\begin{figure}[h]
    \centering
    \includegraphics[width=\linewidth]{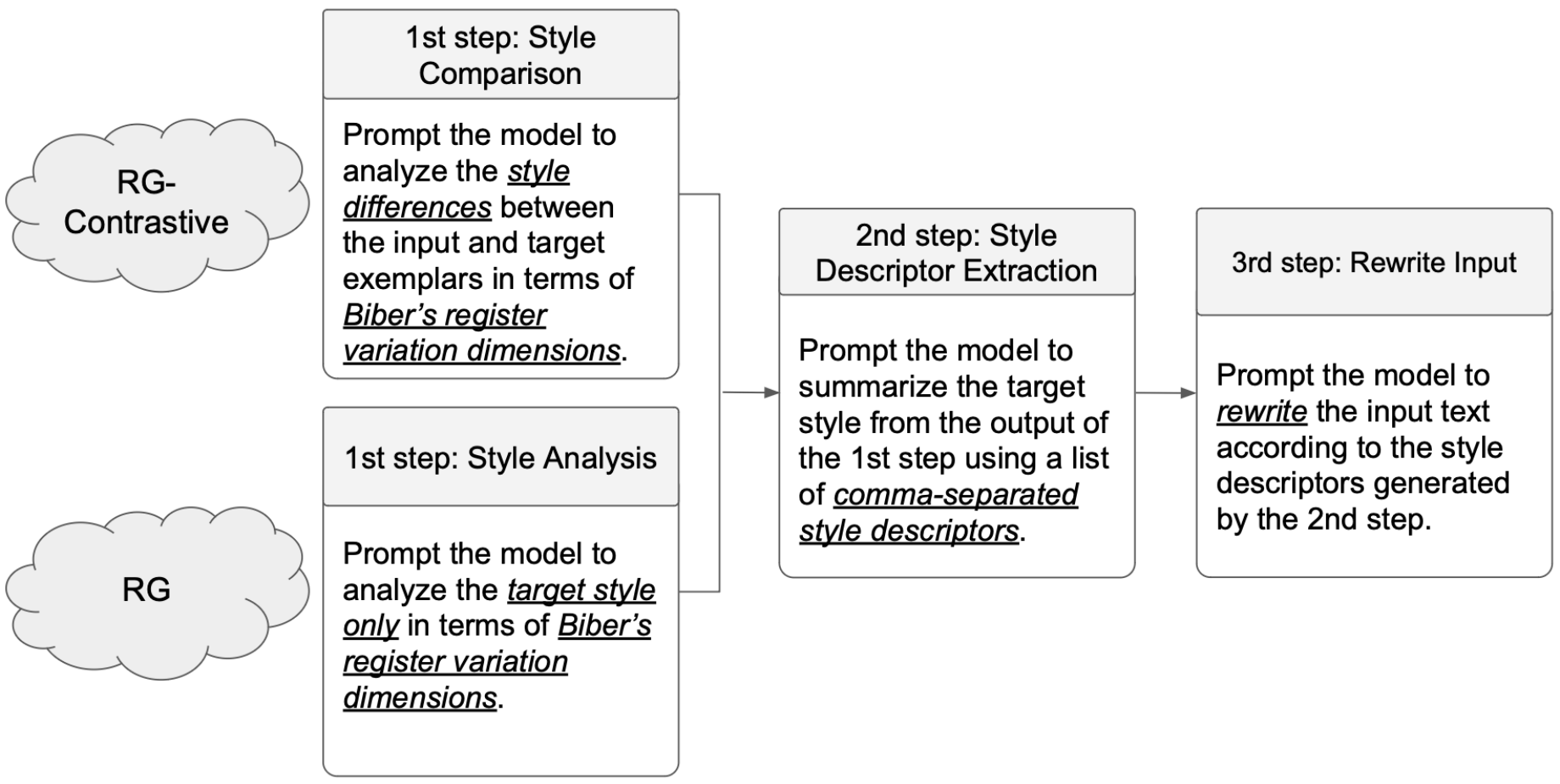}
    \caption{Prompting pipeline for RG-Contrastive and RG, respectively.}
    \label{fig:prompting_pipeline}
\end{figure}
\vspace{-0.5em}

\section{Experiment Setup}

In this section, we describe our experiment setup, including tasks, shared metrics, models and baselines evaluated.

\subsection{Tasks}
\label{tasks}

\subsubsection{Authorship Imitation}
\label{authorship_imitation}

We construct three test sets for the authorship imitation, adopting the method and metrics introduced by ~\citet{styll}.

\paragraph{Datasets.} We sample 15 source authors and 15 target authors from the "test\_queries" and the "test\_targets" split of the Reddit Million User (MUD) dataset ~\citep{MUD} respectively. In the MUD dataset, each author has 16 posts. Following \citet{styll}, we construct three dataset variants: (1) Random: The source and target authors are selected at random. (2) Single: 15 source authors and 15 target authors whose 16 posts all belong to the most common subreddit, ensuring content control by restricting all texts to the same domain. (3) Diverse: 15 source authors and 15 target authors whose 16 posts belong to at least 13 different subreddits, ensuring that any source author or target author writes post in diverse domains. 
See Appendix~\ref{target_construction} for details in target construction.

\paragraph{Task-based metrics.} We use LUAR ~\citep{luar}, an authorship embedding model trained on MUD, to embed input, target and rewritten texts in the authorship space. We calculate the "Away" and "Towards" scores to represent the rewritten-input distance and the rewritten-target distance in the authorship embedding space respectively:

\[
\text{Away} = (1 - \text{Cosine\_similarity}(\text{LUAR}(\text{Rewritten}), \text{LUAR}(\text{Input}))) / 2
\tag{1}
\]
\[
\text{Towards} = (1 + \text{Cosine\_similarity}(\text{LUAR}(\text{Rewritten}), \text{LUAR}(\text{Target}))) / 2
\tag{2}
\]

Following STYLL ~\citep{styll}, we use Mutual Implication Score (MIS) ~\citep{MIS}, a meaning preservation metric based on mutual entailment between two texts, inferred by Natural language Inference (NLI) models.
To improve the reliability of meaning preservation evaluation, we include two additional common metrics, SBERT Similarity \citep{SBERT} and METEOR \citep{METEOR}.

\subsubsection{Formality Transfer}

\paragraph{Datasets.} We use the test split of GYAFC ~\citep{gyafc}, a standard benchmark used for formality transfer evaluation, covering two domains: Entertainment \& Music (EM) and Family \& Relationships (FR). In each domain, we perform formality transfer in two directions: informal to formal (I2F), and formal to informal (F2I). For each direction, we select targets in the desired formality in the train split for rewriting systems to mimic. See Appendix~\ref{target_construction} for details in target construction.

\paragraph{Task-based metrics.} Following previous practices \citep{tinystyler}, we use an off-the-shelf binary formality classifier fine-tuned on GYAFC to evaluate style transfer accuracy \citep{formality_classifier}. We set 0.5 as the decision threshold. We use MIS ~\citep{MIS} to evaluate meaning preservation, following the recommendation in its paper that MIS is particularly successful in a subset of TST tasks including paraphrasing and formality transfer. Again, we include SBERT similarity and METEOR as additional metrics.

\subsubsection{Text Simplification}

\paragraph{Datasets.} 

We evaluate on the test split of Cochrane ~\citep{cochrane}, a paragraph-level simplification task aiming at simplifying medical abstracts into plain-language summaries (PLS). We select PLS texts from the train split of Cochrane as targets for rewriting systems to mimic. See Appendix~\ref{target_construction} for details in target construction.

\paragraph{Task-based metrics.} 

Following previous works \citep{asset, cochrane, KiS}, we use the following metrics. For simplicity: (1) Flesch-Kincaid grade level (FKGL) ~\citep{fk}: A readability test to determine how difficult a passage is by translating into a U.S. school grade level.
(2) Automated Readability Index (ARI) \citep{ari}: Similar to FKGL, ARI is a readability test to gauge the understandability of a text which produces an approximate US grade level needed to comprehend the text.
For content retention: (1) ROUGE \citep{rouge} : A suite of recall-based measures for evaluating content retention in summarization. We report ROUGE-1/2/L scores, which measure unigram, bigram, and longest common subsequence overlap between system output and reference, respectively. (2) BLEU \citep{bleu}: A n-gram, reference-based metric for machine translation that is also often reported for simplification systems \citep{cochrane}. For holistic rewriting quality, we use SARI \citep{SARI}, a metric measuring how well a simplification system performs three key editing operations: keep, delete and add. SARI is found to correlate well with human judgments \citep{SARI, agrawal-carpuat-2024-text}. Lower FKGL and ARI scores indicate better simplicity. ROUGE, BLEU, and SARI range from 0 to 1, with higher scores indicating better quality.

\subsection{Metrics.}
\paragraph{Style Transfer.} In additional to generic style transfer metrics described in individual tasks in Section~\ref{tasks}, we use two stylistic representations to evaluate how accurately the rewritten text mimic the exact target style. (1) StyleCAV ~\citep{styleCAV}: StyleCAV is a style embedding model aiming for "general-purpose" style representation trained with content control on Reddit utterances and tested on Authorship Verification (AV) tasks.
(2) Biber's MDA ~\citep{biber}: Since our method is based on instructing LLMs to rewrite the input text into the target style under the guidance of Biber's MDA, we examine how closely the output aligns with the target in the \textit{actual} Biber's MDA representation space. We follow the practice of \citet{jackgrieve} to perform Biber's MDA on our datasets and system outputs. See Appendix~\ref{biber_mda_implementation} for details on the representation construction and inference procedure. We calculate the "Away" and "Towards" scores based on StyleCAV and Biber's MDA embeddings respectively, as described in Section~\ref{authorship_imitation}.

\paragraph{Meaning Preservation.} See task-based metrics for individual tasks in Section~\ref{tasks}.

\paragraph{Fluency.} Following previous practices \citep{tinystyler, KiP}, we use a gramatical acceptability evaluation model trained on the COLA dataset \citep{COLA, morris2020textattack_cola_model} to evaluate fluency. 

\paragraph{Target Overlap.} To penalize rewriting systems for copying \textit{target content} into output, which is undesirable and may trick style metrics into believing that the output mimics the target style well, we measure the content overlap between system output and target. We report ROUGE-1/2/L scores of the system output based on target as the reference.

\subsection{Models}
We experiment with Llama3.2-3B-Instruct on all the subtasks. We select this model because it is intended for "assistant-like chat and agentic applications like knowledge retrieval and summarization, mobile AI powered writing assistants and query and prompt rewriting" ~\citep{meta-llama-3.2-3b-instruct}. In fact, instruction fine-tuning enables model to follow explicit prompts or instructions and is found to substantially improve zero-shot performance on a diverse range of unseen tasks \citep{RLHF, finetuned_zero-shot_learner, scalinginstructionfinetune, selfinstruct}. The model's relatively small size also makes it well-suited for scenarios with limited computational resources. We aim to explore whether even \textit{smaller} models can benefit from our prompting method. This contrasts ~\citet{recipe}, which employs extremely large (e.g. the 175B GPT-3 ~\citep{gpt3}) models for similar purposes. 

Additionally, we experiment with LLaMA 3.1-8B-Instruct on the authorship imitation subtask to investigate how a moderately larger model responds to our prompting method and whether there are generalizable patterns across models. We do not experiment with larger models from the Llama family (e.g. 70B or above) due to computational limitations. 

We use models from the Huggingface repository and experiment with the default parameters except for "max\_new\_tokens", which we set to 1024 to accommodate longer model outputs. See Table~\ref{tab:model_urls} in Appendix~\ref{model_resources} for access links for the Llama models and model-based metrics mentioned in previous sections.

\subsection{Baselines}

We evaluate our method against the baselines shown in Table~\ref{baselines}. All our experiments are conducted in a zero-shot setting. See Table~\ref{tab:full_prompts} in Appendix~\ref{full_prompts} for full prompts.

\renewcommand{\arraystretch}{1.1}{
\begin{table}[ht]
\centering
\vspace{-1em}
\begin{tabular}{p{2.2cm}p{10cm}}
\hline
\textbf{Method} & \textbf{Description} \\
\hline
\textbf{Copy} & A naive approach that copies the source input text without any modification. \\
\hline
\textbf{Target} & A naive approach that copies the target text as the rewritten text. \\
\hline
\textbf{Gold} & A dummy approach that copies the reference text (if any) as the rewritten text. If multiple reference texts exist, randomly select one as the output. This serves only as an upper bound. \\
\hline
\textbf{Simple} & A simple one-line prompt instructing LLMs to rewrite the input to mimic the authorship style of the target exemplar. \\
\hline
\textbf{STYLL \citep{styll}} & An example-based arbitrary TST method. It first prompts the model to rewrite the input into a neutral style, then to describe the target style using a list of style descriptors, and finally to rewrite the neutral rewrite of input using the style descriptors. \\
\hline
\end{tabular}
\caption{Overview of the experimental baselines.}
\label{baselines}
\end{table}
}

\section{Results}
\begin{figure}[ht]
    \centering
    \includegraphics[width=\linewidth]{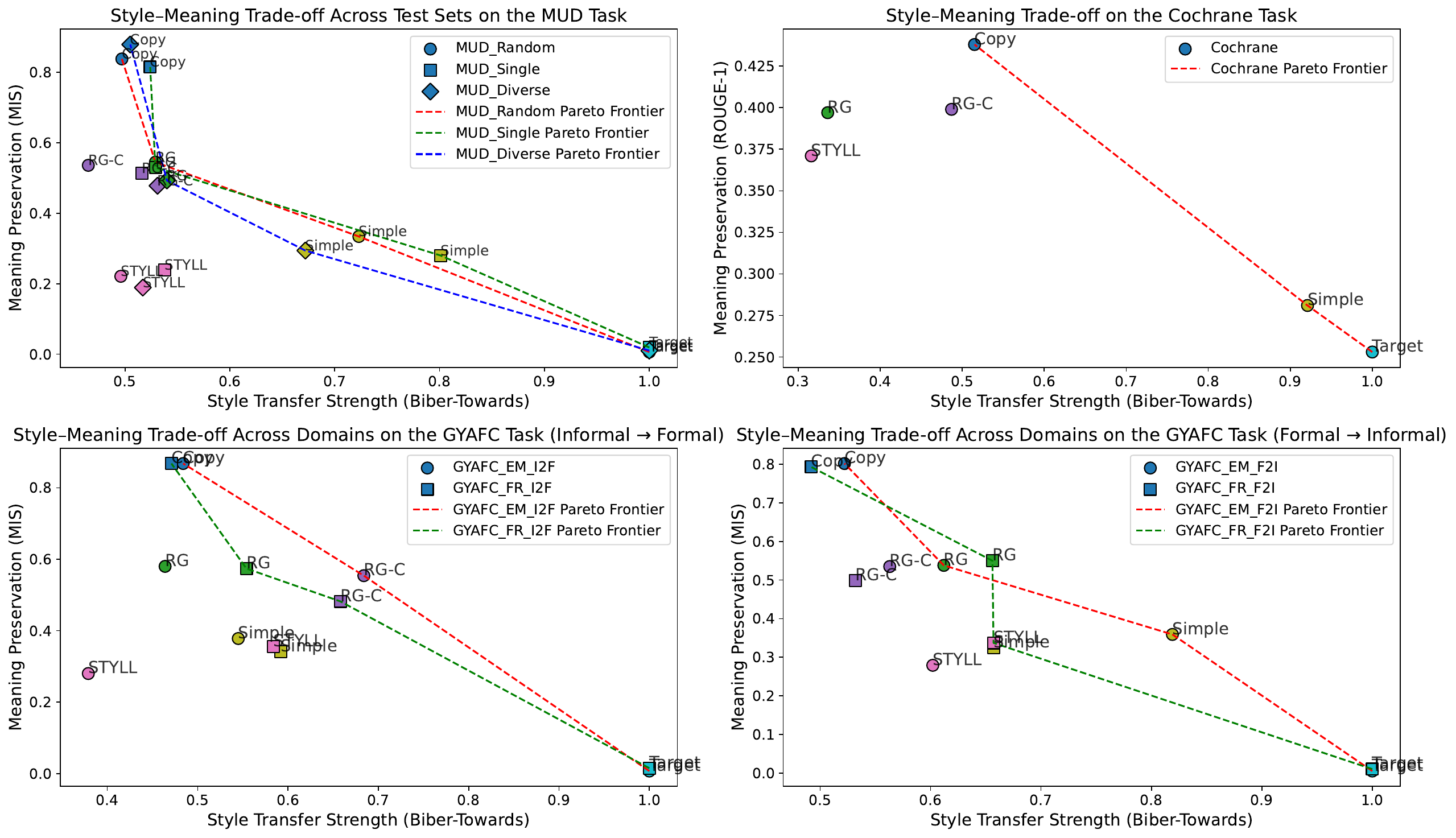}
    \caption{Style–meaning trade-offs across tasks: MUD, Cochrane, and GYAFC (both directions), using Llama-3.2-3B-Instruct. Pareto frontiers identify systems that achieve optimal trade-offs for each task. RG-C: RG-Contrastive.}
    \label{fig:pareto_frontiers}
\end{figure}

\begin{table}[ht]
\centering
\centering
\vspace{-1em}
\resizebox{\textwidth}{!}{
\begin{tabular}{l|ccc|cccc|c}
\toprule
\textbf{System} & \multicolumn{3}{c|}{\textbf{MUD (Overlap Rouge-1 $\downarrow$)}} & \multicolumn{4}{c|}{\textbf{GYAFC (Overlap Rouge-1 $\downarrow$)}} & \textbf{Cochrane} \\
& Random & Single & Diverse & EM\_I2F & EM\_F2I & FR\_I2F & FR\_F2I & \textbf{(Overlap Rouge-1 $\downarrow$)} \\
\midrule
Copy   & 0.075 & 0.051 & 0.070 & 0.234 & 0.299 & 0.141 & 0.088 & 0.234 \\
Target & 1.000 & 1.000 & 1.000 & 1.000 & 1.000 & 1.000 & 1.000 & 1.000 \\
Gold (dummy)   & --    & --    & --    & 0.253 & 0.087 & 0.089 & 0.095 & 0.253 \\
\midrule
Simple & 0.343 & 0.426 & 0.382 & 0.864 & 0.310 & 0.127 & 0.104 & 0.864 \\
STYLL  & 0.145 & 0.119 & 0.148 & 0.239 & 0.056 & 0.099 & 0.090 & 0.239 \\
\rowcolor{cyan!20}
RG-Contrastive   & \textbf{0.107} & \textbf{0.070} & \textbf{0.108} & 0.251 & \textbf{0.046} & 0.104 & 0.094 & 0.251 \\
\rowcolor{cyan!20}
RG     & 0.110 & 0.077 & 0.113 & \textbf{0.234} & 0.090 & \textbf{0.095} & \textbf{0.086} & \textbf{0.234} \\
\midrule
\textbf{System} & \multicolumn{3}{c|}{\textbf{MUD (COLA $\uparrow$)}} & \multicolumn{4}{c|}{\textbf{GYAFC (COLA $\uparrow$)}} & \textbf{Cochrane} \\
& Random & Single & Diverse & EM\_I2F & EM\_F2I & FR\_I2F & FR\_F2I & \textbf{(COLA $\uparrow$)} \\
\midrule
Copy   & 0.783 & 0.679 & 0.754 & 0.746 & 0.933 & 0.790 & 0.921 & 0.969 \\
Target & 0.067 & 0.133 & 0.067 & 0.253 & 0.000 & 0.520 & 0.042 & 0.965 \\
Gold (dummy)   & --    & --    & --    & 0.925 & 0.743 & 0.944 & 0.822 & 0.967 \\
\midrule
Simple & 0.537 & 0.446 & 0.532 & 0.727 & 0.517 & 0.916 & 0.745 & 0.967 \\
STYLL  & \textbf{0.982} & \textbf{0.956} & \textbf{0.978} & 0.970 & \textbf{0.951} & 0.989 & \textbf{0.987} & \textbf{1.000} \\
\rowcolor{cyan!20}
RG-Contrastive   & 0.960 & 0.951 & 0.960 & \textbf{0.983} & 0.944 & \textbf{0.990} & 0.986 & \textbf{1.000} \\
\rowcolor{cyan!20}
RG     & 0.949 & 0.923 & 0.952 & 0.929 & 0.939 & 0.977 & 0.972 & 0.998 \\
\bottomrule
\end{tabular}
}
\caption{Target Overlap (measured by Rouge-1, lower is better) and grammatical acceptability (measured by COLA, higher is better) scores for rewriting systems across MUD, GYAFC, and Cochrane tasks, using Llama-3.2-3B-Instruct. \textbf{Bold} values indicate the best scores among non-naive systems (Simple, STYLL, RG-Contrastive, RG).}
\label{tab:utility_scores}
\end{table}

For arbitrary TST by example, to study the trade-off between style transfer strength and meaning preservation, which is the core challenge of this task, we plot the Pareto frontiers of rewriting systems across style transfer strength (x-axis, measured by "Towards" using Biber's MDA representation) and meaning preservation (y-axis, measured by MIS on MUD and GYAFC, and by Rouge-1 on Cochrane), evaluated on Llama3.2-3B-Instruct (see Fig~\ref{fig:pareto_frontiers}). For each task, the Pareto frontier shows the set of systems for which no other achieves better performance in both objectives simultaneously. We observe that naive baselines "copy" and "target" are always on the Pareto frontier, as expected, as "copy" and "target" achieves the best meaning preservation and style imitation respectively. 

Among the evaluated systems, our method consistently demonstrates strong performance across tasks. On the MUD authorship imitation task, RG is on the Pareto frontier across all data splits, achieving a good balance between style transfer strength and meaning preservation, while RG-Contrastive performs slightly behind. On GYAFC, trend varies by transfer direction. In the I2F direction, RG-Contrastive sits on the Pareto frontier across both EM and FR domains, achieving much stronger style transfer than RG. While in the F2I direction, it is the opposite - RG sits on the Pareto frontier across domains and provides better style transfer strength than RG-Contrastive. The performance distinction potentially suggests different use cases of whether including the contrasting mechanism or not. We hypothesize that the "flipping trend" is because the "formal" targets in the GYAFC dataset are not formal in an \textit{absolute} sense but \textit{relative} to the input text. In I2F experiments, we observe that RG and STYLL are inclined to generate descriptors such as "informal" and "causal", misleading the model into the wrong direction, but RG-Contrastive usually gets it right. On the contrary, the "informal" targets in the GYAFC dataset are mostly informal in an \textit{absolute} sense (with lots of slangs, abbreviations, etc.), so RG itself is sufficient while the contrasting mechanism may confound the model. On Cochrane, RGs miss the Pareto frontier, with RG-Contrastive performs better than RG but beaten by "copy" - it moves the composite style of the input text slightly away from the target. This may be because the input and target texts in Cochrane have prominent baseline similarities, i.e. the "technical" register as the input and target are original and simplified versions of medical abstract respectively, so small perturbations to the input style may inadvertently move it further away from the target style. For example, RG and STYLL tend to arrive at the wrong direction by determining that the target style is "technical", while RG-Contrastive sometimes decides that the target style is "informal" compared to the input text, which points out the right direction but may be an overshoot, reflecting the intrinsic challenge of pinpointing the accurate target position in the style space, especially when the target style occupies an intermediate position rather than a distinct stylistic endpoint. 

In contrast, STYLL, a state-of-the-art zero-/few-shot example-based arbitrary TST method, is consistently a suboptimal choice across tasks (miss Pareto Frontier and often dominated by RG-Contrastive/RG except on the GYAFC\_FR\_F2I task). It achieves style transfer strength that is at best comparable to RGs and often worse, while consistently lagging behind in meaning preservation by a large margin. Surprisingly, "Simple" frequently sits on the Pareto frontier and beats sophisticated methods (STYLL, RGs) in style transfer strength. However, this may not be out of genuine style transfer but instead be inflated by target content copying into rewrites. As seen from Table~\ref{tab:utility_scores}, which summarizes the utility scores of the systems across tasks, "Simple" frequently gets $3 \sim 5$ times higher Overlap Rouge-1 score than RGs and STYLL, and scores even over 0.8 on Cochrane, indicating a non-trivial higher level of target content copying which can defeat the original purpose of style transfer (especially on Cochrane). "Simple" also has much lower COLA scores than RGs and STYLL on MUD and 3 out of 4 GYAFC splits, which to some extent resemble "Targets" whose low linguistic acceptability is expected because each target of MUD and GYAFC is a concatenation of multiple semantically and logically unrelated texts - potentially due to copying behavior.

\begin{table*}[ht]
\centering
\vspace{-1em}
\resizebox{\textwidth}{!}{
\begin{tabular}{l|cc|cc|cc|cc||ccc|c}
\toprule
\textbf{System} & \multicolumn{8}{c||}{\textbf{GYAFC}} & \multicolumn{4}{c}{\textbf{Cochrane}} \\
\cmidrule(lr){2-9} \cmidrule(lr){10-13}
 & \multicolumn{2}{c|}{EM\_I2F} & \multicolumn{2}{c|}{EM\_F2I} & \multicolumn{2}{c|}{FR\_I2F} & \multicolumn{2}{c||}{FR\_F2I} & FKGL $\downarrow$ & ARI $\downarrow$ & SARI $\uparrow$ & Rouge-1 $\uparrow$ \\
 & Acc $\uparrow$ & MIS $\uparrow$ & Acc $\uparrow$ & MIS $\uparrow$ & Acc $\uparrow$ & MIS $\uparrow$ & Acc $\uparrow$ & MIS $\uparrow$ & & & & \\
\midrule
Copy & 0.064 & 0.868 & 0.024 & 0.802 & 0.072 & 0.868 & 0.046 & 0.793 & 12.81 & 15.26 & 0.418 & 0.438 \\
Target & 0.999 & 0.007 & 0.887 & 0.005 & 0.996 & 0.015 & 0.845 & 0.010 & 11.68 & 13.86 & 0.346 & 0.253 \\
Gold (dummy) & 0.921 & 0.877 & 0.830 & 0.787 & 0.937 & 0.877 & 0.850 & 0.780 & 11.49 & 13.66 & 0.982 & 1.000 \\
\midrule
Simple & 0.476 & 0.378 & \textbf{0.869} & 0.359 & 0.553 & 0.341 & \textbf{0.820} & 0.325 & 11.76 & 13.95 & 0.353 & 0.281 \\
STYLL & 0.554 & 0.280 & 0.533 & 0.279 & 0.641 & 0.355 & 0.500 & 0.337 & 13.61 & 15.53 & 0.382 & 0.371 \\
\rowcolor{cyan!20}
RG-Contrastive & \textbf{0.886} & 0.554 & 0.482 & 0.535 & \textbf{0.899} & 0.482 & 0.396 & 0.499 & \textbf{11.47} & \textbf{13.58} & \textbf{0.390} & \textbf{0.399} \\
\rowcolor{cyan!20}
RG & 0.347 & \textbf{0.580} & 0.707 & \textbf{0.538} & 0.423 & \textbf{0.574} & 0.647 & \textbf{0.550} & 14.55 & 17.16 & 0.374 & 0.397 \\
\bottomrule
\end{tabular}
}
\caption{Evaluation results of generic style transfer on the GYAFC and Cochrane tasks, using Llama-3.2-3B-Instruct. \textbf{Left:} GYAFC is evaluated with accuracy ($\uparrow$) and meaning preservation (MIS $\uparrow$). \textbf{Right:} Cochrane is evaluated with readability (FKGL, ARI $\downarrow$), editing-quality (SARI $\downarrow$) and meaning preservation (Rouge-1 $\uparrow$). \textbf{Bold} values indicate the best scores among non-naive systems (Simple, STYLL, RG-Contrastive, RG).}
\vspace{-1em}
\label{tab:generic_ST}
\end{table*}

Table~\ref{tab:generic_ST} summarizes the style transfer strength in terms of the major dimension of the downstream task, if any (formality for GYAFC, simplicity for Cochrane). On GYAFC (I2F), RG-contrastive achieves the best style transfer accuracy, agreeing with its clear advantage in imitating the target style on this task. On GYAFC (F2I), "Simple" leads and RG gets the 2nd place in terms of accuracy, ahead of STYLL by a large margin. Notably, both RG variants show a clear edge towards STYLL and "Simple" in meaning preservation indicated by MIS (e.g. on EM\_I2F, RG-Contrastive and "Simple" achieve an MIS score of 0.554 and 0.378 respectively). On Cochrane, RG-Contrastive leads across all four metrics (FKGL, ARI, SARI, Rouge-1). Unlike the slight underperformance in terms of accurately replicating the target style on this task, RG-Contrastive serves the downstream goal very well. It \textit{indeed} moves the input text toward the "simple" end of the simplicity spectrum and even achieves an overall simplicity better than that of "Gold" (indicated by FKGL and ARI). Overall, it suggests that RG-Contrastive/RG is highly effective at picking up major style dimensions of the target text and performing style transfer across these dimensions compared to other methods such as STYLL. Although RG-Contrastive/RG may not always capture the infinite composite nuances of a target style, which is intrinsically hard, it can serve practical downstream goals well especially when there are prominent major style transfer dimensions.

The other metrics show similar trends, and we report full experimental results (Llama3.2-3B-Instruct results on MUD, GYAFC and Cochrane, and Llama3.1-8B-Instruct results on MUD) in Appendix~\ref{complete_results}. Llama3.1-8B-Instruct on MUD shows similar trends to those observed on Llama3.2-3B-Instruct across rewriting systems: RGs achieve a great balance between style transfer strength and meaning preservation and show a clear advantage in meaning preservation, with RG often dominating STYLL in both objectives across MUD splits. "Simple", again, leads in "Towards" scores, but exhibits substantially higher target overlap, indicating a higher degree of non-genuine rewriting through target copying and suffers from low linguistic acceptability if "Target" displays the same "issue" (e.g. by concatenating multiple texts). This indicates the the patterns we observed on Llama3.2-3B-Instruct is not reserved to a single model but can be generalizable to other models. Compared to Llama3.2-3B-Instruct, Llama3.1-8B-Instruct generally shows improved meaning preservation (indicated by MIS, etc.) across splits, similar to dropped style transfer strength (indicated by Towards-Biber, etc.) on the Random and Single split, and similar to improved style transfer strength on the Diverse split. This indicates that a slightly larger model can provide benefits in meaning preservation, not necessarily in style transfer strength, but may have an edge in cross-domain style imitation, where the input and target are from different fields.

\section{Qualitative Analysis}
\paragraph{Qualitative Examples.} To get a qualitative understanding of the behavior of the rewriting systems, we take a look at outputs generated by them. Table~\ref{tab:style_transfer_examples} in Appendix~\ref{qualitative_examples} shows a few (input, target, outputs) examples on the MUD task. The systems rewrite the input \textit{"Verratti is practically untouchable. He's signing an extension every year or so and PSG won't sell for even a €100m."} towards different targets ("...Aaaaanndd you are all on a list...", "...Jesus Christ, Cesaro...", etc.). In the given examples, compared to the input, targets are generally more informal and conversational than the input. RGs successfully capture this shift, producing outputs with stylistic markers such as verbal fillers (“oh man”, “i mean”), lowercase ("i mean", "no way"), and expressions like “nope” and “lol”. In terms of meaning, RGs accurately retain key meaning including "untouchable", "sign an extension yearly", "PSG would not let Verratti go in any way" accross the targets. STYLL successfully captures the shift towards a more informal style but introduces greater meaning distortion—for example, by adding content not present in the input ("locking down new deals", "bread and butter of the team", "legend", "glue", etc.). In fact, STYLL adds details that are plausible or potentially inferable from the input context, but are not explicitly present in the original text. Thus, RG-based method is the more appropriate choice for tasks with strict meaning preservation requirements. "Simple" shows a milder tonal shift, occasionally using conversational and informal language in its outputs, but it also occasionally introduces content from the target ("Log in you ... ones!"), which is an undesired behavior.

\begin{figure}[!t]
    \centering
    \vspace{-1em}
    \includegraphics[width=\linewidth]{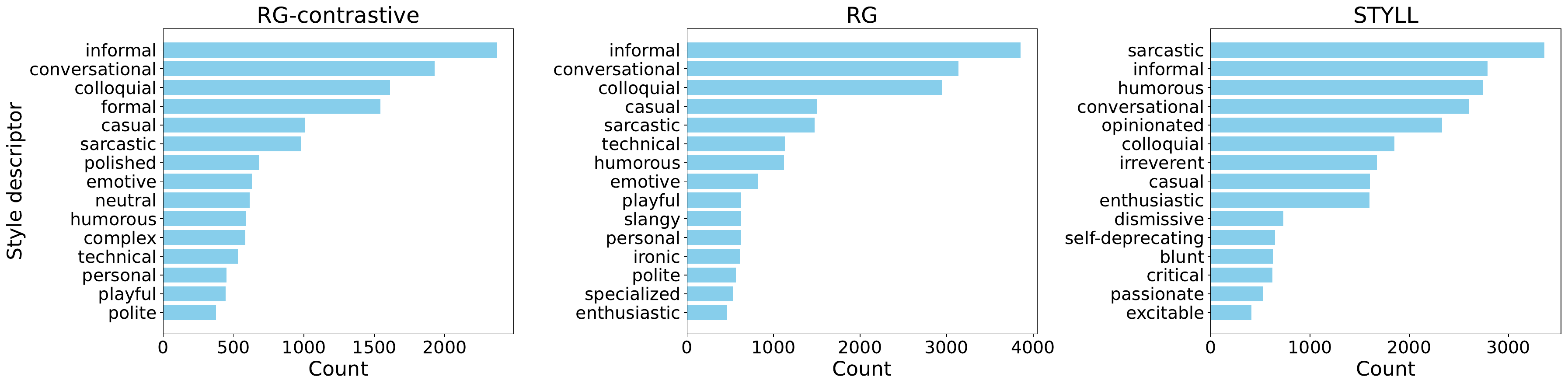}
    \caption{Top 15 style descriptors by frequency by generated rewriting system (RG-Contrastive, RG, STYLL) on the MUD\_Random task.}
    \label{fig:style_descriptors}
\end{figure}

\paragraph{Style Descriptor Analysis.} For both RGs and STYLL, the style descriptors generated during the intermediate step serve as a key characterization of the target exemplar and determine the direction of style transfer. While STYLL leaves LLMs to interpret "style" in an open-ended way, RG-based prompting instructs the model to interpret "style" using Biber's register analysis as guidance. To understand whether different ways of interpreting "style" affect the nature of the descriptors produced, we conduct a statistical analysis comparing the distributions of style descriptors generated by the methods (RG-Contrastive, RG, STYLL). Figure~\ref{fig:style_descriptors} presents the top 15 most frequent style descriptors generated by each system on the MUD\_Random dataset. We observe that the top 3 most frequent descriptors generated by both RGs are "informal", "conversational" and "colloquial", which aligns with the informal and conversational nature of MUD, a dataset constructed from Reddit. In contrast, STYLL’s top 3 descriptors are "sarcastic", "informal", and "humorous", with two of them differing from RGs' top 3. Beyond this, RGs and STYLL also generate descriptors unique to their respective interpretations of "style". For example, STYLL's top 15 leaderboard includes descriptors such as "opinionated", "irreverent", "dismissive", "self-deprecating", etc., which are not among the top 15 most frequent descriptors of RGs. On the contrary, RGs’ top 15 includes "polished", "emotive", "playful", "technical", "polite", etc., which are not among the top 15 of STYLL. Overall, STYLL tends to produce more affective, tone-oriented style descriptors, which may signal shifts in tone or intent and thus are more likely to alter the original meaning. In contrast, RGs' style descriptors are more restricted to the register space, guiding the model towards stylistic adjustments with minimal impact on semantics.

\section{Conclusion} 
In this paper, we introduce a prompting method based on register analysis to guide LLMs in example-based arbitrary TST tasks. Rather than relying on open-ended interpretations of “style,” our method instructs the model to reason about stylistic variation within the space of register. Experimental results across multiple style transfer tasks show that our method achieves enhanced rewriting quality, especially in meaning preservation. Furthermore, qualitative analysis shows that our method produces style descriptors that are more closely aligned with register, with minimal changes in intent or tone, thereby reducing the risk of inadvertent meaning alteration. Overall, our prompting strategy strengthens stylistic control while maintaining higher semantic accuracy than existing methods.

\section*{Acknowledgments}
We thank Zoey (Dayeon) Ki, HyoJung Han, Kartik Ravisankar in the CLIP Lab of UMD for their constructive feedback.

This research is supported in part by the Office of the Director of National Intelligence (ODNI), Intelligence Advanced Research Projects Activity (IARPA), via the HIATUS Program contract \#2022-22072200006. The views and conclusions contained herein are those of the authors and should not be interpreted as necessarily representing the official policies, either expressed or implied, of ODNI, IARPA, or the U.S. Government. The U.S. Government is authorized to reproduce and distribute reprints for governmental purposes notwithstanding any copyright annotation therein.

\section*{Ethics Statement}
We acknowledge the potential risks associated with the technique described in this paper, including possible misuse by malicious actors for impersonation attacks targeting authorship attribution (AA) and authorship verification (AV) systems, as well as the propagation of misinformation. However, we believe that the benefits of our work outweigh these risks. Our technique enhances style transfer in low-resource settings, enables applications such as personalized writing assistance and privacy preservation for online users, and serves as a testbed for evaluating the adversarial robustness of AA and AV systems. 

To promote the benefits of this work while mitigating potential harms, we advocate for practices of responsible use. For example, style-based generation may be encouraged for creative, educational or research purposes, while being appropriately regulated in identity-sensitive or official contexts. Additionally, disclosure in usage can be adopted to ensure transparency and accountability.

\bibliography{colm2025_conference}
\bibliographystyle{colm2025_conference}

\appendix
\newpage
\section{Full Prompts}
\label{full_prompts}
\begin{table*}[h]
    \centering
    \renewcommand{\arraystretch}{1.4}
    \setlength{\tabcolsep}{12pt}
    \rowcolors{2}{gray!15}{white}
    \begin{tabular}{p{1cm} p{12cm}}
        \toprule
        \textbf{\#} & \textbf{Simple} \\
        \midrule
        \rowcolor{gray!15}
        1 & Here is the target text \texttt{[target text]} Rewrite \texttt{[input text]} into the authorship style of the target text. Strictly output only the rewritten text without any other content. \\
        \bottomrule
    \end{tabular}
    \begin{tabular}{p{1cm} p{12cm}}
        \textbf{\#} & \textbf{STYLL} \\
        \midrule
        1 & Source text: Passage: \texttt{[source text]} Paraphrase the passage in a simple neutral style. \\
        2 & Passage: \texttt{[target text]} List some adjectives, comma-separated, that describe the writing style of the author of this passage. Strictly output only the style descriptors without any other content. \\
        3 & Here is a text: [neural paraphrase] Here is a rewrite of the text that is more [style descriptors]. Strictly output only the rewritten text without any other content. \\
        \bottomrule
    \end{tabular}
    \begin{tabular}{p{1cm} p{12cm}}
        \rowcolor{white}
        \textbf{\#} & \textbf{RG-Contrastive} \\
        \midrule
        \rowcolor{gray!15}
        1 & Source text: \texttt{[source text]} Target text: \texttt{[target text]} How does the target text differ from the source text in authorship style in terms of dimensions of register variation according to Douglas Biber? \\
        \rowcolor{white}
        2 & Style comparisons: \texttt{[style comparisons]} List some adjectives, comma-separated, that describe the writing style of the author of the target text. Strictly output only the style descriptors without any other content. \\
        \rowcolor{gray!15}
        3 & Here is a text: \texttt{[source text]} Rewrite the text to be more \texttt{[style descriptors]}. Strictly output only the rewritten text without any other content. \\
        \bottomrule
    \end{tabular}
    \begin{tabular}{p{1cm} p{12cm}}
        \textbf{\#} & \textbf{RG} \\
        \midrule
        1 & Passage: \texttt{[target text]} Analyze the authorship style of this passage in terms of dimensions of register variation according to Douglas Biber. \\
        2 & Style analysis: \texttt{[style analysis]} List some adjectives, comma-separated, that describe the writing style of the author of the target text. Strictly output only the style descriptors without any other content. \\
        3 & Here is a text: \texttt{[source text]} Rewrite the text to be more \texttt{[style descriptors]}. Strictly output only the rewritten text without any other content. \\
        \bottomrule
    \end{tabular}
    \caption{Full prompts used in experiments for Simple, STYLL, RG-Contrastive and RG, respectively. In the table, [neural paraphrase / style comparisons / style analysis] is the model's output from the 1st step and [style descriptors] is the model's output from the 2nd step. For STYLL, we adopt the specific prompts outlined in \citet{styll}, but with minor tweaks to adapt to the zero-shot setting in our experiments.}
    \label{tab:full_prompts}
\end{table*}

\newpage
\section{Target Construction}
\label{target_construction}

\paragraph{Authorship Imitation.} During inference-time, each \textit{individual} input text from any of the source authors (each source author has 16 such texts) is paired a with each of the target authors for the model to perform authorship style transfer (ST). In total, there are $16 \text{ (number of individual texts per source author)} \times 15 \text{ (number of source authors)} \times 15 \text{ (number of target authors)} = 3600$ such pairings. During each ST corresponding to each pairing, the 16 texts from the target author are concatenated together as the target exemplar whose style the model is tasked to rewrite the input text into.

\paragraph{Formality Transfer.} To avoid accidentally exposing the "gold answer" to the rewriting systems, we select targets from the train split of the GYAFC dataset. Texts in GYAFC are of sentence-level and thus may be too short to contain enough linguistic information to inform the LLM. Hence, for each input text, we concatenate $K=16$ texts randomly selected from the target selection pool to form a paragraph-level target exemplar, following our MUD evaluation set construction practices for authorship imitation. Targets have the same domain and opposite formality as their corresponding input text. For example, for an formal input text in the "EM" domain, 16 sentence-level texts are randomly selected from the EM-train-informal split of GYAFC to form a paragraph-level target to inform style during model's inference time.

\paragraph{Text Simplification.} 
To avoid accidentally exposing the "gold answer" to the rewriting systems, we randomly select plain-language summary texts from the train split of the Cochrane dataset as target. Unlike GYAFC, texts in Cochrane are of paragraph-level, so each of them suffices in length to serve as a single target without the need of concatenation.

\section{Model Resources}
\label{model_resources}

Table~\ref{tab:model_urls} provides links to the pretrained models used in our experiments. All models are publicly available and can be accessed through the Hugging Face Model Hub.

\begin{table}[h]
\centering
 \resizebox{\textwidth}{!}{
\begin{tabular}{ll}
\toprule
\textbf{Model} & \textbf{URL} \\
\midrule
Llama3.2-3B-Instruct & \url{https://huggingface.co/meta-llama/Llama-3.2-3B-Instruct} \\
Llama3.1-8B-Instruct & \url{https://huggingface.co/meta-llama/Llama-3.1-8B-Instruct} \\
LUAR & \url{https://huggingface.co/rrivera1849/LUAR-MUD} \\
SBERT & \url{https://huggingface.co/sentence-transformers/all-MiniLM-L6-v2} \\
Formality classifier (DeBERTa) & \url{https://huggingface.co/s-nlp/deberta-large-formality-ranker} \\
StyleCAV & \url{https://huggingface.co/AnnaWegmann/Style-Embedding} \\
COLA & \url{https://huggingface.co/textattack/roberta-base-CoLA} \\
\bottomrule
\end{tabular}
}
\caption{Links to pretrained models used in our experiments.}
\label{tab:model_urls}
\end{table}

\newpage
\section{Implementation Details of Biber's MDA Style Representation}
\label{biber_mda_implementation}

\begin{figure}[h]
    \centering
    \includegraphics[width=\linewidth]{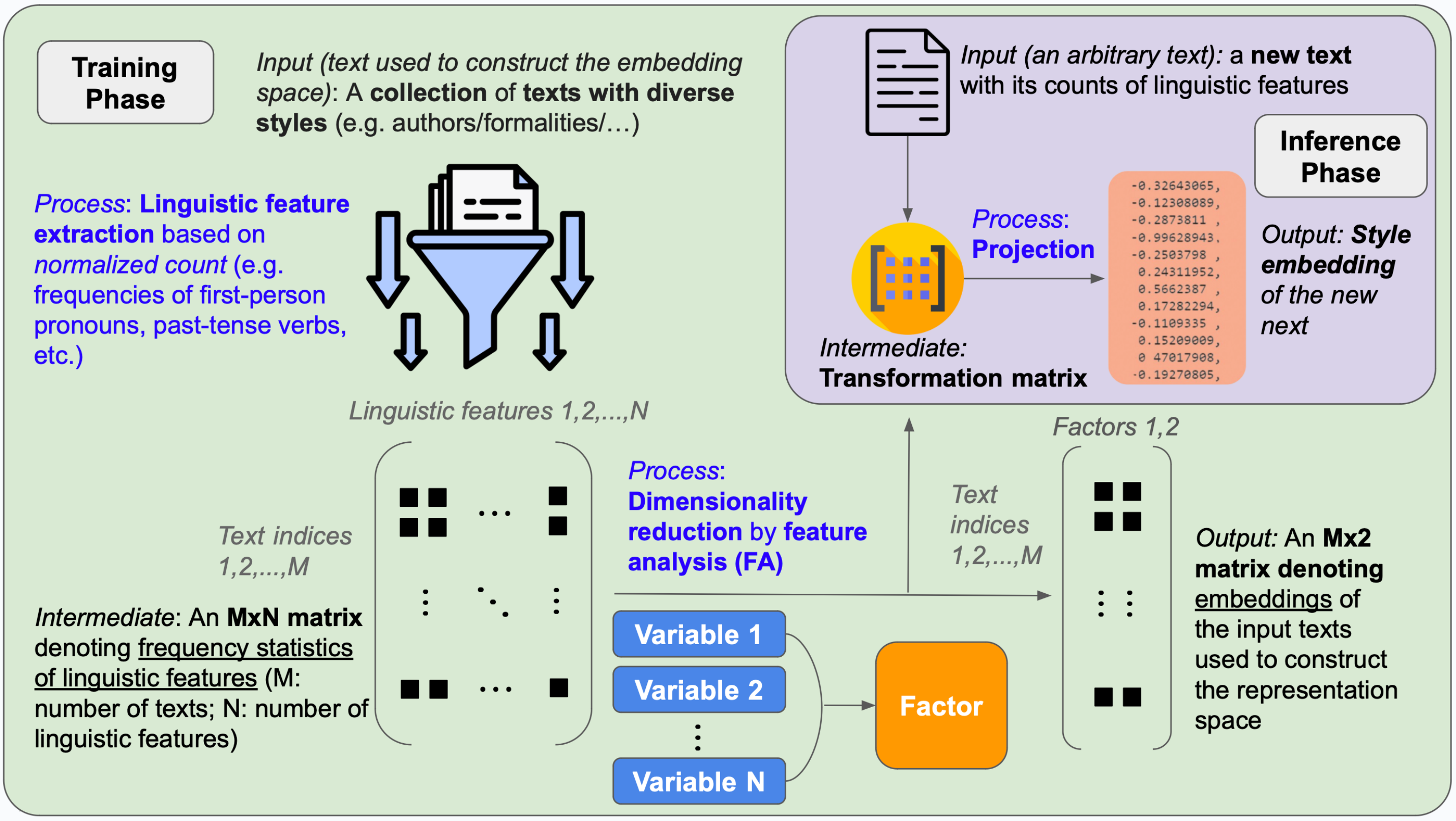}
    \caption{Illustration of the procedure of building Biber’s MDA representation from a linguistic corpus and using it to make inference on a new text following the practice of \citet{jackgrieve}.}
    \label{fig:biber_mda_procedure}
\end{figure}

The train and inference phrase of Biber's MDA representation are shown in Fig~\ref{fig:biber_mda_procedure}. We train one Biber's MDA representation per task (MUD, GYAFC, Cochrane). The training corpus for each task encompasses a diverse range of styles for each task (authorship for MUD, formality for GYAFC, and simplicity for Cochrane). The data splits constituting the training corpus for each task are specified as follows: \\
\begin{itemize}
    \item MUD: validation-queries, validation-targets. \\
    \item GYAFC: EM-train-formal, EM-train-informal, FR-train-formal, FR-train-informal. \\
    \item Cochrane: train-original, train-simplified.
\end{itemize}

\newpage
\section{Complete Results}
\label{complete_results}

\subsection{Authorship Imitation}
\label{mud_results}
\begin{table}[h]
    \centering
    \Huge
    \resizebox{\textwidth}{!}{ %
    \begin{tabular}{llcccccccccccccc}
        \specialrule{3pt}{0pt}{0pt}
        \multirow{2}{*}{\textbf{Random}} & \multirow{2}{*}{\textbf{System}} & \textbf{Away} & \textbf{Towards} & \multirow{2}{*}{\textbf{MIS}} & \multirow{2}{*}{\textbf{Sbert}} & \multirow{2}{*}{\textbf{Meteor}} & \multirow{2}{*}{\textbf{COLA}} & \textbf{Away} & \textbf{Towards} & \textbf{Away} & \textbf{Towards} & \textbf{Overlap} & \textbf{Overlap} & \textbf{Overlap} \\
         & & \textbf{(LUAR)} & \textbf{(LUAR)} & & & & & \textbf{(StyleCAV)} & \textbf{(StyleCAV)} & \textbf{(Biber)} & \textbf{(Biber)} & \textbf{Rouge-1 ($\downarrow$)} & \textbf{Rouge-2 ($\downarrow$)} & \textbf{Rouge-L ($\downarrow$)} \\
        \midrule
        \multirow{2}{*}{\textbf{Naive}} 
        & Copy & 0.000 & 0.617 & 0.838 & 1.000 & 0.994 & 0.783 & 0.000 & 0.544 & 0.000 & 0.497 & 0.075 & 0.005 & 0.045 \\
         & Target & 0.383 & 1.000 & 0.006 & 0.065 & 0.117 & 0.067 & 0.456 & 1.000 & 0.503 & 1.000 & 1.000 & 1.000 & 1.000 \\
        \midrule
        \multirow{2}{*}{\textbf{Llama3.2-3B}} 
         & Simple & 0.269 & \textbf{0.757} & 0.334 & 0.475 & \textbf{0.471} & 0.537 & 0.349 & \textbf{0.731} & 0.340 & \textbf{0.723} & 0.343 & 0.279 & 0.314 \\
         & STYLL & \textbf{0.330} & 0.612 & 0.221 & 0.490 & 0.222 & \textbf{0.982} & 0.368 & 0.522 & \textbf{0.457} & 0.496 & 0.145 & 0.018 & 0.078 \\
         \textbf{-Instruct} & RG-C & 0.278 & 0.604 & 0.536 & 0.647 & 0.338 & 0.960 & \textbf{0.369} & 0.473 & 0.425 & 0.465 & \textbf{0.107} & \textbf{0.009} & \textbf{0.061} \\
         & RG & 0.272 & 0.607 & \textbf{0.545} & \textbf{0.661} & 0.365 & 0.949 & 0.333 & 0.530 & 0.362 & 0.529 & 0.110 & 0.010 & 0.062 \\
        \midrule
        \multirow{2}{*}{\textbf{Llama3.1-8B}}
         & Simple & 0.250 & \textbf{0.732} & 0.451 & 0.580 & \textbf{0.565} & 0.602 & 0.340 & \textbf{0.731} & 0.292 & \textbf{0.672} & 0.280 & 0.211 & 0.247 \\
         & STYLL & \textbf{0.324} & 0.632 & 0.261 & 0.548 & 0.235 & \textbf{0.981} & \textbf{0.369} & 0.534 & \textbf{0.443} & 0.491 & 0.178 & 0.021 & 0.089 \\
         \textbf{-Instruct} & RG-C & 0.282 & 0.611 & 0.578 & 0.654 & 0.346 & 0.969 & 0.342 & 0.495 & 0.434 & 0.483 & \textbf{0.140} & \textbf{0.013} & \textbf{0.073} \\
         & RG & 0.273 & 0.610 & \textbf{0.602} & \textbf{0.668} & 0.372 & 0.963 & 0.325 & 0.522 & 0.406 & 0.499 & 0.141 & 0.014 & 0.074 \\
        \specialrule{3pt}{0pt}{0pt}
        
        \\\\\\
        
         \specialrule{3pt}{0pt}{0pt}
        \multirow{2}{*}{\textbf{Single}} & \multirow{2}{*}{\textbf{System}} & \textbf{Away} & \textbf{Towards} & \multirow{2}{*}{\textbf{MIS}} & \multirow{2}{*}{\textbf{Sbert}} & \multirow{2}{*}{\textbf{Meteor}} & \multirow{2}{*}{\textbf{COLA}} & \textbf{Away} & \textbf{Towards} & \textbf{Away} & \textbf{Towards} & \textbf{Overlap} & \textbf{Overlap} & \textbf{Overlap} \\
         & & \textbf{(LUAR)} & \textbf{(LUAR)} & & & & & \textbf{(StyleCAV)} & \textbf{(StyleCAV)} & \textbf{(Biber)} & \textbf{(Biber)} & \textbf{Rouge-1 ($\downarrow$)} & \textbf{Rouge-2 ($\downarrow$)} & \textbf{Rouge-L ($\downarrow$)} \\
        \midrule
        \multirow{2}{*}{\textbf{Naive}} 
        & Copy & 0.000 & 0.635 & 0.815 & 1.000 & 0.992 & 0.679 & -0.000 & 0.555 & 0.000 & 0.524 & 0.051 & 0.004 & 0.033 \\
         & Target & 0.365 & 1.000 & 0.019 & 0.111 & 0.091 & 0.133 & 0.445 & 1.000 & 0.476 & 1.000 & 1.000 & 1.000 & 1.000 \\
        \midrule
        \multirow{2}{*}{\textbf{Llama3.2-3B}}  
         & Simple & 0.265 & \textbf{0.793} & 0.280 & 0.414 & \textbf{0.411} & 0.446 & 0.346 & \textbf{0.787} & 0.349 & \textbf{0.801} & 0.426 & 0.388 & 0.411 \\
         & STYLL & \textbf{0.315} & 0.618 & 0.239 & 0.450 & 0.209 & \textbf{0.956} & 0.394 & 0.503 & 0.426 & 0.538 & 0.119 & 0.026 & 0.075 \\
         \textbf{-Instruct} & RG-C & 0.263 & 0.611 & 0.514 & 0.618 & 0.334 & 0.951 & \textbf{0.406} & 0.457 & \textbf{0.429} & 0.516 & \textbf{0.070} & \textbf{0.005} & \textbf{0.045} \\
         & RG & 0.252 & 0.612 & \textbf{0.530} & \textbf{0.654} & 0.390 & 0.923 & 0.324 & 0.524 & 0.329 & 0.529 & 0.077 & 0.008 & 0.048 \\
        \midrule
        \multirow{2}{*}{\textbf{Llama3.1-8B}}   & Simple & 0.244 & \textbf{0.759} & 0.404 & 0.534 & \textbf{0.505} & 0.496 & 0.361 & \textbf{0.804} & 0.315 & \textbf{0.770} & 0.327 & 0.287 & 0.311 \\
         & STYLL & \textbf{0.313} & 0.625 & 0.284 & 0.507 & 0.226 & \textbf{0.958} & \textbf{0.379} & 0.554 & \textbf{0.438} & 0.500 & 0.133 & 0.014 & 0.070 \\
         \textbf{-Instruct} & RG-C & 0.264 & 0.613 & 0.532 & 0.626 & 0.341 & 0.946 & 0.377 & 0.476 & 0.434 & 0.483 & \textbf{0.091} & \textbf{0.008} & \textbf{0.054} \\
         & RG & 0.259 & 0.610 & \textbf{0.541} & \textbf{0.646} & 0.368 & 0.935 & 0.325 & 0.529 & 0.383 & 0.510 & 0.099 & 0.010 & 0.057 \\
        \specialrule{3pt}{0pt}{0pt}

        \\\\\\

        \specialrule{3pt}{0pt}{0pt}
        \multirow{2}{*}{\textbf{Diverse}} & \multirow{2}{*}{\textbf{System}} & \textbf{Away} & \textbf{Towards} & \multirow{2}{*}{\textbf{MIS}} & \multirow{2}{*}{\textbf{Sbert}} & \multirow{2}{*}{\textbf{Meteor}} & \multirow{2}{*}{\textbf{COLA}} & \textbf{Away} & \textbf{Towards} & \textbf{Away} & \textbf{Towards} & \textbf{Overlap} & \textbf{Overlap} & \textbf{Overlap} \\
         & & \textbf{(LUAR)} & \textbf{(LUAR)} & & & & & \textbf{(StyleCAV)} & \textbf{(StyleCAV)} & \textbf{(Biber)} & \textbf{(Biber)} & \textbf{Rouge-1 ($\downarrow$)} & \textbf{Rouge-2 ($\downarrow$)} & \textbf{Rouge-L ($\downarrow$)} \\
        \midrule
        
        \multirow{2}{*}{\textbf{Naive}} 
        & Copy & 0.000 & 0.613 & 0.879 & 1.000 & 0.989 & 0.754 & -0.000 & 0.567 & 0.000 & 0.505 & 0.070 & 0.005 & 0.041 \\
         & Target & 0.387 & 1.000 & 0.010 & 0.086 & 0.113 & 0.067 & 0.433 & 1.000 & 0.495 & 1.000 & 1.000 & 1.000 & 1.000 \\
        \midrule
        \multirow{2}{*}{\textbf{Llama3.2-3B}}  
         & Simple & 0.286 & \textbf{0.755} & 0.294 & 0.457 & \textbf{0.446} & 0.532 & 0.344 & \textbf{0.719} & 0.332 & \textbf{0.672} & 0.382 & 0.313 & 0.352 \\
         & STYLL & \textbf{0.340} & 0.597 & 0.189 & 0.490 & 0.221 & \textbf{0.978} & \textbf{0.380} & 0.532 & \textbf{0.421} & 0.517 & 0.148 & 0.015 & 0.076 \\
         \textbf{-Instruct} & RG-C & 0.286 & 0.594 & 0.478 & 0.631 & 0.336 & 0.960 & 0.378 & 0.495 & 0.400 & 0.531 & \textbf{0.108} & \textbf{0.009} & \textbf{0.061} \\
         & RG & 0.285 & 0.598 & \textbf{0.493} & \textbf{0.649} & 0.364 & 0.952 & 0.339 & 0.533 & 0.348 & 0.540 & 0.113 & 0.011 & 0.062 \\
        \midrule
        \multirow{2}{*}{\textbf{Llama3.1-8B}}
        & Simple & 0.275 & \textbf{0.753} & 0.367 & 0.520 & \textbf{0.509} & 0.524 & 0.339 & \textbf{0.740} & 0.298 & \textbf{0.678} & 0.380 & 0.320 & 0.354 \\
         & STYLL & \textbf{0.342} & 0.621 & 0.200 & 0.527 & 0.209 & \textbf{0.968} & \textbf{0.389} & 0.532 & \textbf{0.423} & 0.551 & 0.206 & 0.020 & 0.096 \\
         \textbf{-Instruct}  & RG-C & 0.296 & 0.602 & 0.520 & 0.637 & 0.338 & 0.965 & 0.368 & 0.510 & 0.392 & 0.540 & \textbf{0.149} & \textbf{0.014} & \textbf{0.076} \\
         & RG & 0.294 & 0.602 & \textbf{0.531} & \textbf{0.651} & 0.347 & 0.964 & 0.352 & 0.531 & 0.376 & 0.553 & 0.155 & 0.015 & 0.078 \\
        \specialrule{3pt}{0pt}{0pt}
    \end{tabular}
    }
    \caption{Evaluation results of different rewriting systems on the MUD task, using Llama-3.2-3B-Instruct and Llama-3.1-8B-Instruct, respectively. \textbf{Bold} values indicate the best scores among non-naive systems (Simple, STYLL, RG-C, RG). RG-C: RG-Contrastive. }
    \label{tab:mud_results}
\end{table}

\newpage
\subsection{Formality Transfer}
\label{gyafc_results}
\begin{table}[h]
    \centering
    \Huge
    \resizebox{\textwidth}{!}{ %
    \begin{tabular}{llcccccccccccccccc}
\specialrule{3pt}{0pt}{0pt}
\multirow{2}{*}{\textbf{Setting}} & \multirow{2}{*}{\textbf{System}} & \textbf{Acc} & \multirow{2}{*}{\textbf{MIS}} & \multirow{2}{*}{\textbf{Sbert}} & \multirow{2}{*}{\textbf{Meteor}} & \multirow{2}{*}{\textbf{COLA}} & \textbf{Away} & \textbf{Towards} & \textbf{Away} & \textbf{Towards} & \textbf{Overlap} & \textbf{Overlap} & \textbf{Overlap} \\
 & & \textbf{(DeBERTa)} & & & & & \textbf{(StyleCAV)} & \textbf{(StyleCAV)} & \textbf{(Biber)} & \textbf{(Biber)} & \textbf{Rouge-1 ($\downarrow$)} & \textbf{Rouge-2 ($\downarrow$)} & \textbf{Rouge-L ($\downarrow$)} \\
\midrule

\multirow{7}{*}{\textbf{EM\_I2F}} 
& Copy & 0.064 & 0.868 & 0.824 & 0.738 & 0.746 & 0.000 & 0.554 & 0.000 & 0.484 & 0.059 & 0.002 & 0.054 \\
 & Target & 0.999 & 0.007 & 0.100 & 0.137 & 0.253 & 0.446 & 1.000 & 0.516 & 1.000 & 1.000 & 1.000 & 1.000 \\
 & Gold & 0.921 & 0.877 & 0.876 & 0.999 & 0.925 & 0.439 & 0.633 & 0.267 & 0.598 & 0.071 & 0.004 & 0.065 \\
 \cmidrule(lr){2-14}
 & Simple & 0.476 & 0.378 & 0.461 & 0.432 & 0.727 & 0.385 & \textbf{0.682} & 0.280 & 0.545 & 0.299 & 0.201 & 0.265 \\
 & STYLL & 0.554 & 0.280 & 0.490 & 0.292 & 0.970 & 0.488 & 0.603 & 0.361 & 0.379 & 0.141 & 0.010 & 0.080 \\
 & RG-C & \textbf{0.886} & 0.554 & 0.599 & 0.402 & \textbf{0.983} & \textbf{0.544} & 0.535 & \textbf{0.429} & \textbf{0.684} & \textbf{0.087} & \textbf{0.006} & 0.060 \\
 & RG & 0.347 & \textbf{0.580} & \textbf{0.630} & \textbf{0.443} & 0.929 & 0.406 & 0.584 & 0.307 & 0.464 & 0.088 & \textbf{0.006} & \textbf{0.058} \\

\specialrule{3pt}{0pt}{0pt}

\\\\\\

\specialrule{3pt}{0pt}{0pt}

\multirow{2}{*}{\textbf{Setting}} & \multirow{2}{*}{\textbf{System}} & \textbf{Acc} & \multirow{2}{*}{\textbf{MIS}} & \multirow{2}{*}{\textbf{Sbert}} & \multirow{2}{*}{\textbf{Meteor}} & \multirow{2}{*}{\textbf{COLA}} & \textbf{Away} & \textbf{Towards} & \textbf{Away} & \textbf{Towards} & \textbf{Overlap} & \textbf{Overlap} & \textbf{Overlap} \\
 & & \textbf{(DeBERTa)} & & & & & \textbf{(StyleCAV)} & \textbf{(StyleCAV)} & \textbf{(Biber)} & \textbf{(Biber)} & \textbf{Rouge-1 ($\downarrow$)} & \textbf{Rouge-2 ($\downarrow$)} & \textbf{Rouge-L ($\downarrow$)} \\
\midrule

\multirow{7}{*}{\textbf{EM\_F2I}} 
 & Copy & 0.024 & 0.802 & 0.738 & 0.595 & 0.933 & 0.000 & 0.338 & 0.000 & 0.522 & 0.056 & 0.001 & 0.053 \\
 & Target & 0.887 & 0.005 & 0.110 & 0.128 & 0.000 & 0.662 & 1.000 & 0.478 & 1.000 & 1.000 & 1.000 & 1.000 \\
 & Gold & 0.830 & 0.787 & 0.796 & 0.999 & 0.743 & 0.461 & 0.654 & 0.334 & 0.666 & 0.046 & 0.002 & 0.043 \\
 \cmidrule(lr){2-14}
 & Simple & \textbf{0.869} & 0.359 & 0.436 & 0.330 & 0.517 & \textbf{0.534} & \textbf{0.749} & 0.388 & \textbf{0.819} & 0.310 & 0.236 & 0.286 \\
 & STYLL & 0.533 & 0.279 & 0.473 & 0.301 & \textbf{0.951} & 0.366 & 0.378 & \textbf{0.452} & 0.602 & 0.127 & 0.009 & 0.072 \\
 
  & RG-C & 0.482 & 0.535 & 0.584 & 0.398 & 0.944 & 0.326 & 0.379 & 0.357 & 0.563 & \textbf{0.089} & \textbf{0.006} & \textbf{0.057} \\
  & RG & 0.707 & \textbf{0.538} & \textbf{0.598} & \textbf{0.424} & 0.939 & 0.319 & 0.387 & 0.358 & 0.612 & 0.090 & 0.007 & \textbf{0.057} \\

\specialrule{3pt}{0pt}{0pt}

\\\\\\

\specialrule{3pt}{0pt}{0pt}

\multirow{2}{*}{\textbf{Setting}} & \multirow{2}{*}{\textbf{System}} & \textbf{Acc} & \multirow{2}{*}{\textbf{MIS}} & \multirow{2}{*}{\textbf{Sbert}} & \multirow{2}{*}{\textbf{Meteor}} & \multirow{2}{*}{\textbf{COLA}} & \textbf{Away} & \textbf{Towards} & \textbf{Away} & \textbf{Towards} & \textbf{Overlap} & \textbf{Overlap} & \textbf{Overlap} \\
 & & \textbf{(DeBERTa)} & & & & & \textbf{(StyleCAV)} & \textbf{(StyleCAV)} & \textbf{(Biber)} & \textbf{(Biber)} & \textbf{Rouge-1 ($\downarrow$)} & \textbf{Rouge-2 ($\downarrow$)} & \textbf{Rouge-L ($\downarrow$)} \\
\midrule

\multirow{7}{*}{\textbf{FR\_I2F}} 
 & Copy & 0.072 & 0.868 & 0.797 & 0.748 & 0.790 & 0.000 & 0.550 & 0.000 & 0.471 & 0.070 & 0.003 & 0.063 \\
 & Target & 0.996 & 0.015 & 0.093 & 0.155 & 0.520 & 0.450 & 1.000 & 0.529 & 1.000 & 1.000 & 1.000 & 1.000 \\
 & Gold & 0.937 & 0.877 & 0.848 & 1.000 & 0.944 & 0.437 & 0.620 & 0.249 & 0.628 & 0.081 & 0.005 & 0.072 \\
 \cmidrule(lr){2-14}
 & Simple & 0.553 & 0.341 & 0.421 & 0.400 & 0.916 & 0.438 & \textbf{0.658} & 0.337 & 0.592 & 0.214 & 0.118 & 0.179 \\
 & STYLL & 0.641 & 0.355 & 0.434 & 0.299 & 0.989 & 0.501 & 0.604 & 0.371 & 0.584 & 0.099 & 0.008 & 0.063 \\
 & RG-C & \textbf{0.899} & 0.482 & 0.475 & 0.314 & \textbf{0.990} & \textbf{0.567} & 0.501 & \textbf{0.480} & \textbf{0.658} & 0.104 & \textbf{0.007} & 0.067 \\
 & RG & 0.423 & \textbf{0.574} & \textbf{0.552} & \textbf{0.420} & 0.977 & 0.449 & 0.577 & 0.323 & 0.554 & \textbf{0.095} & \textbf{0.007} & \textbf{0.061} \\

\specialrule{3pt}{0pt}{0pt}

\\\\\\

\specialrule{3pt}{0pt}{0pt}

\multirow{2}{*}{\textbf{Setting}} & \multirow{2}{*}{\textbf{System}} & \textbf{Acc} & \multirow{2}{*}{\textbf{MIS}} & \multirow{2}{*}{\textbf{Sbert}} & \multirow{2}{*}{\textbf{Meteor}} & \multirow{2}{*}{\textbf{COLA}} & \textbf{Away} & \textbf{Towards} & \textbf{Away} & \textbf{Towards} & \textbf{Overlap} & \textbf{Overlap} & \textbf{Overlap} \\
 & & \textbf{(DeBERTa)} & & & & & \textbf{(StyleCAV)} & \textbf{(StyleCAV)} & \textbf{(Biber)} & \textbf{(Biber)} & \textbf{Rouge-1 ($\downarrow$)} & \textbf{Rouge-2 ($\downarrow$)} & \textbf{Rouge-L ($\downarrow$)} \\
\midrule

\multirow{7}{*}{\textbf{FR\_I2F}}
 & Copy & 0.046 & 0.793 & 0.683 & 0.590 & 0.921 & 0.000 & 0.360 & 0.000 & 0.492 & 0.055 & 0.004 & 0.042 \\
 & Target & 0.845 & 0.010 & 0.094 & 0.144 & 0.042 & 0.640 & 1.000 & 0.508 & 1.000 & 1.000 & 1.000 & 1.000 \\
 & Gold & 0.850 & 0.780 & 0.751 & 0.999 & 0.822 & 0.471 & 0.678 & 0.322 & 0.618 & 0.050 & 0.003 & 0.037 \\
 \cmidrule(lr){2-14}
 & Simple & \textbf{0.820} & 0.325 & 0.394 & 0.312 & 0.745 & \textbf{0.464} & \textbf{0.666} & \textbf{0.457} & \textbf{0.657} & 0.213 & 0.127 & 0.182 \\
 & STYLL & 0.500 & 0.337 & 0.406 & 0.294 & \textbf{0.987} & 0.297 & 0.335 & 0.374 & \textbf{0.657} & 0.090 & \textbf{0.007} & 0.057 \\
 & RG-C & 0.396 & 0.499 & 0.487 & 0.352 & 0.986 & 0.316 & 0.332 & 0.347 & 0.532 & 0.094 & \textbf{0.007} & 0.060 \\
 & RG & 0.647 & \textbf{0.550} & \textbf{0.515} & \textbf{0.404} & 0.972 & 0.305 & 0.365 & 0.336 & 0.656 & \textbf{0.086} & \textbf{0.007} & \textbf{0.056} \\
\specialrule{3pt}{0pt}{0pt}

\end{tabular}
    }
    \caption{Evaluation results of different rewriting systems on the GYAFC task, across EM and FR domains for both informal-to-formal (I2F) and formal-to-informal (F2I) directions, using Llama-3.2-3B-Instruct. \textbf{Bold} values indicate the best scores among non-naive systems (Simple, STYLL, RG-C, RG). RG-C: RG-Contrastive. RG-C: RG-Contrastive. }
    \label{tab:gyafc_results}
\end{table}

\subsection{Cochrane}
\label{cochrane_results}
\begin{table}[h]
    \centering
    \Huge
    \resizebox{\textwidth}{!}{%
    \begin{tabular}{llccccccccccccccccccc}
\specialrule{3pt}{0pt}{0pt}

\multirow{2}{*}{\textbf{System}} & \multirow{2}{*}{\textbf{FKGL ($\downarrow$)}} & \multirow{2}{*}{\textbf{ARI ($\downarrow$)}} & \multirow{2}{*}{\textbf{Rouge-1}} & \multirow{2}{*}{\textbf{Rouge-2}} & \multirow{2}{*}{\textbf{Rouge-L}} & \multirow{2}{*}{\textbf{BLEU}} & \multirow{2}{*}{\textbf{SARI}} & \multirow{2}{*}{\textbf{COLA}} & \textbf{Away} & \textbf{Towards} & \textbf{Away} & \textbf{Towards} & \textbf{Overlap} & \textbf{Overlap} & \textbf{Overlap} \\
& & & & & & & & & \textbf{(StyleCAV)} & \textbf{(StyleCAV)} & \textbf{(Biber)} & \textbf{(Biber)} & \textbf{Rouge-1 ($\downarrow$)} & \textbf{Rouge-2 ($\downarrow$)} & \textbf{Rouge-L ($\downarrow$)} \\
\midrule

Copy & 12.81 & 15.26 & 0.438 & 0.196 & 0.250 & 0.132 & 0.418 & 0.969 & 0.000 & 0.903 & 0.000 & 0.515 & 0.234 & 0.028 & 0.113 \\
Target & 11.68 & 13.86 & 0.253 & 0.031 & 0.125 & 0.005 & 0.346 & 0.965 & 0.097 & 1.000 & 0.485 & 1.000 & 1.000 & 1.000 & 1.000 \\
Gold & 11.49 & 13.66 & 1.000 & 1.000 & 1.000 & 1.000 & 0.982 & 0.967 & 0.071 & 0.907 & 0.393 & 0.610 & 0.253 & 0.031 & 0.125 \\
\midrule
 
Simple & 11.76 & 13.95 & 0.281 & 0.054 & 0.143 & 0.022 & 0.353 & 0.967 & 0.086 & \textbf{0.974} & 0.425 & \textbf{0.921} & 0.864 & 0.832 & 0.851 \\
STYLL & 13.61 & 15.53 & 0.371 & 0.102 & 0.200 & 0.041 & 0.382 & \textbf{1.000} & 0.096 & 0.889 & \textbf{0.598} & 0.316 & 0.239 & \textbf{0.025} & 0.129 \\
RG-C & \textbf{11.47} & \textbf{13.58} & \textbf{0.399} & 0.126 & 0.207 & 0.068 & \textbf{0.390} & \textbf{1.000} & \textbf{0.120} & 0.860 & 0.551 & 0.487 & 0.251 & 0.028 & 0.125 \\
RG & 14.55 & 17.16 & 0.397 & \textbf{0.141} & \textbf{0.213} & \textbf{0.082} & 0.374 & 0.998 & 0.088 & 0.873 & 0.365 & 0.336 & \textbf{0.234} & 0.027 & \textbf{0.118} \\
 \specialrule{3pt}{0pt}{0pt}
 
\end{tabular}
     }
    \caption{
    Evaluation results of different rewriting systems on the Cochrane dataset, using Llama-3.2-3B-Instruct. \textbf{Bold} values indicate the best scores among non-naive systems (Simple, STYLL, RG-C, RG). RG-C: RG-Contrastive. 
    }
    \label{tab:cochrane_results}
\end{table}

\newpage
\section{Qualitative examples}
\label{qualitative_examples}
\begin{table*}[h]
\centering
\large
\resizebox{\textwidth}{!}{%
\begin{tabular}{p{3.8cm} p{3.8cm} p{4.8cm} p{4.8cm} p{4.8cm} p{4.8cm}}
\toprule
\textbf{Target} & \textbf{Input} & \textbf{RG-Contrastive} & \textbf{RG} & \textbf{STYLL} & \textbf{Simple} \\
\midrule
"He didn't do any of that"
\newline
"Aaaaanndd you are all on a list." 
\newline 
"Oh for fucks sake You are brilliant \textgreater Hearing ``DADDDDYYYYY!!'' " 
\newline …13 more examples (concatenated together)
& Verratti is practically untouchable. He's signing an extension every year or so and PSG won't sell for even a €100m.
& oh man, verratti is basically untouchable, no way psg would sell him for anything less than a gazillion euros, and even then they'd probably just give him a massive extension to keep him happy, i mean, the guy's basically a god on the pitch
& Dude, Verratti is basically untouchable - he's always signing a new deal and PSG would never even consider selling him for a hundred million euros, let alone that.
& Verratti's not going anywhere, lol. He's always locking down new deals and PSG is like ``nah, we're not selling him for more than a sweet, sweet €100m.''
& Verratti is practically untouchable, he's signing an extension every year or so \& PSG won't sell for even a €100m. \\

\midrule

"Log in you damned ones!" \newline "Loving the Japanese commentary team." \newline "Jesus Christ, Cesaro!" \newline "Theatrics, dude." \newline …12 more examples (concatenated together)
& Verratti is practically untouchable. He's signing an extension every year or so and PSG won't sell for even a €100m.
& Verratti is basically untouchable, like he's signing extension after extension every year or so and PSG is all ``nope, not selling him for a single euro.''
& Verratti is basically untouchable – he's signing a new deal every year or so and PSG is like, ``no way, he's staying.'' They wouldn't even sell him for a cool €100m.
& Oh man, you think PSG is just gonna let Verratti walk out the door? Please, they're basically married to the guy. He's like the bread and butter of the team, and they're not about to let him go for a measly €100m. I mean, come on, they're already paying him to sit on the bench, let alone sell him for a fraction of his worth. He's basically a legend, and they're not about to let him become a legend elsewhere.
& Log in you damned ones! Verratti's like untouchable, dude. He's signing extensions left and right, PSG's gonna be stuck with him for ages. They wouldn't sell him for a €100m, that's just not gonna happen. \\

\midrule

"I actually rate my coffee houses on two actors: the music and the coffee. " \newline "I'm humble as fuck." \newline "Jesus, when did we start talking about French food?",  \newline …13 more examples (concatenated together)
& Verratti is practically untouchable. He's signing an extension every year or so and PSG won't sell for even a €100m.
& Verratti is practically untouchable. He's signing an extension every year or so and PSG won't sell for even a €100m.
I mean, come on, Verratti's basically untouchable, right? Every year or so, he's signing an extension deal and PSG's all like ``no way, we're not selling him for a cool €100m.'' I mean, seriously, what even is the point of having a €100m clause if you're not gonna use it, lol?
& Verratti is basically untouchable, he's always signing a new deal and PSG won't even consider selling him for a hundred million euros.
& Verratti's like glue, man — nobody can get rid of him. He's always signing a new contract, so PSG is basically stuck with him. And if they do decide to sell him, they're only gonna get €100m, max.
& Verratti's a total god, son. He's signing an extension every year or so and PSG ain't gonna let him go for even a €100m. \\
\midrule

"Bioshock has been confirmed for a long time though. The Bioshocks were confirmed to be coming soon when they launched backward compatibility I believe. " \newline …15 more examples (concatenated together)
& Verratti is practically untouchable. He's signing an extension every year or so and PSG won't sell for even a €100m.
& so i think Verratti's basically unmoveable — always seems to be signing a new deal every year or so and PSG's like, ``nah, we're not letting him go for less than €100m.''
& Dude, Verratti is basically untouchable — every year or so he's signing a new deal and PSG is all like ``no way, we're not selling him for less than €100m.''
& ``Come on, you think PSG is gonna just dump Verratti like a hot potato because he signs a new contract? Give me a break. The guy's a genius on the field and they can't even be bothered to try and get rid of him. They're just gonna hold on to him for dear life because they know they can't sell him for a pretty penny.''
& Verratti is basically unmovable, he's always signing some kind of extension and PSG isn't gonna let him go for even a €100m.
\\
\midrule

"This analogy doesn't really work for Jim, per se" \newline "And her lopsided breasts." \newline "Hmm, cum causes neurological problems? Is that why you have to ``medicate'' your head?" \newline …13 more examples (concatenated together)
& Verratti is practically untouchable. He's signing an extension every year or so and PSG won't sell for even a €100m.
& It's a well-documented fact that Verratti has an ironclad contract, with an annual extension that's almost as frequent as a Parisian café's espresso refills. Meanwhile, PSG's valuation remains stubbornly resistant to even the most astronomical offers, standing firm like a Gallic fortress at €100m.
& OMG, you guys, Verratti is basically untouchable, lol. Every year he's signing a new deal and PSG is like `no way, we're not selling him for even a cent'. I mean, come on, a €100m fee? Please, they'd rather lose him for free than sell him to some other team.
& Oh boy, PSG is totally desperate to keep Verratti, and he's just thrilled to be stuck with them. Like, he's literally counting down the days until his next contract extension, and the club is just over the moon that he won't be selling him to anyone for a gazillion bucks.
& Verratti's practically untouchable, dude. He's signing an extension every year or so and PSG's gonna be like ``no way, we're not selling him for even a €100m, he's our golden boy.
\\
\bottomrule
\end{tabular}
}
\caption{Example outputs generated by different rewriting systems conditioned on target-style exemplars.}
\label{tab:style_transfer_examples}
\end{table*}

\end{document}